\documentclass[runningheads]{llncs}

\usepackage[export]{adjustbox}
\usepackage{graphicx}
\graphicspath{ {./figs/} }
\usepackage{subcaption} 

\usepackage[noadjust]{cite} 

\usepackage{amsmath}
\usepackage{amsfonts}  
\usepackage[cal=boondox]{mathalfa} 

\usepackage{algorithm}
\usepackage{algorithmic}

\usepackage[table]{xcolor} 
\usepackage{tabularx}
\usepackage{multirow} 
\usepackage{adjustbox} 
\usepackage{booktabs} 

\usepackage[misc]{ifsym} 
\usepackage{bbding} 

\usepackage[colorlinks=true, linkcolor=magenta, urlcolor=blue, citecolor=cyan, bookmarks=true, bookmarksopen, pagebackref=true]{hyperref} 
\title{Graph Generative Models Evaluation with Masked Autoencoder}
\author{
 Chengen Wang \Envelope\inst{1} \and
 Murat Kantarcioglu\inst{2}
}

\institute{
University of Texas at Dallas, Dallas TX 75080, USA\\
\email{chengen.wang@utdallas.edu} \and
Virginia Tech, Blacksburg, VA 24061, USA\\
\email{muratk@vt.edu}
}

\begin{document}
\maketitle

\begin{abstract}
In recent years, numerous graph generative models (GGMs) have been proposed. However, evaluating these models remains a considerable challenge, primarily due to the difficulty in extracting meaningful graph features that accurately represent real-world graphs. The traditional evaluation techniques, which rely on graph statistical properties like node degree distribution, clustering coefficients, or Laplacian spectrum, overlook node features and lack scalability. There are newly proposed deep learning-based methods employing graph random neural networks or contrastive learning to extract graph features, demonstrating superior performance compared to traditional statistical methods, but their experimental results also demonstrate that these methods do not always working well across different metrics. Although there are overlaps among these metrics, they are generally not interchangeable, each evaluating generative models from a different perspective. In this paper, we propose a novel method that leverages graph masked autoencoders to effectively extract graph features for GGM evaluations. We conduct extensive experiments on graphs and empirically demonstrate that our method can be more reliable and effective than previously proposed methods across a number of GGM evaluation metrics, such as ``Fréchet Distance (FD)" and ``MMD Linear". However, no single method stands out consistently across all metrics and datasets. Therefore, this study also aims to raise awareness of the significance and challenges associated with GGM evaluation techniques, especially in light of recent advances in generative models.
Our code is available at \href{https://github.com/chengenw/ggmEval}{https://github.com/chengenw/ggmEval}

\end{abstract}

\section{Introduction}\label{sec:intro}
Graph generative models (GGMs) have important applications across different domains, including biology, chemistry, engineering and social networks. Recent advances in graph generative models underscore the need for robust metrics to evaluate the synthetic graphs in comparison to the real graphs \cite{o'bray2022evaluation}. Unlike images, human visual perception is hardly applicable to graphs due to their complex structures. In addition, commonly-used feature extractors, like Inception network \cite{inception_score(IS)}, are not readily available for the graph domain.

Traditional graph evaluation methods rely on representations derived from general graph topology, such as node degree distributions, clustering coefficient distributions, orbit counts and the Laplacian spectrum. Although these statistics capture important structural properties, they neglect features associated with individual nodes, which limits their ability to provide expressive representations of graphs. Furthermore, these methods often face scalability challenges.

\cite{o'bray2022evaluation} highlights issues with commonly-used evaluation methods, which may lead to inconsistent rankings across different settings. This underscores the urgent need to develop new GGM evaluation techniques.

The key challenge in evaluating GGM lies in effectively extracting graph representations. With faithful graph representations, the extracted features or embeddings can be used as input for standard graph evaluation metrics, such as the Fréchet Distance \cite{FID}, and Precision \& Recall \cite{recall}, which we will discuss in Section \ref{sec:pre}. 

Recently \cite{ggm_thompson} proposed using graph random neural networks to extract graph features, demonstrating the advantages of the deep learning-based methods. Building on this, \cite{ggm_shirzad2022evaluating} introduced a self-supervised contrastive learning method to extract more faithful graph representations, generally outperforming the previous technique.

Graph masked autoencoders (GMAE) \cite{graphmae_mask,maskgae_JintangLi} represent another self-supervised learning method that can achieve performance on par with or even surpass, contrastive learning-based methods. Graph masked autoencoder works by recovering masked node features or edges. The experimental results from \cite{graphmae_mask} demonstrate GMAE can \emph{match or exceed} contrastive learning method, raising the question of whether graph masked autoencoder can more effectively extract graph representations, potentially leading to improved GGM evaluation techniques.

In this work, we propose to leverage graph masked autoencoders to assess the fidelity and diversity of graph generative models. Through extensive experiments, we demonstrate the effectiveness and advantages of our proposed method compared to existing evaluation approaches across multiple metrics.

Our main contributions are as follows:
\begin{enumerate}
    \item We propose leveraging a graph masked autoencoder to extract graph representations, thereby offering a more effective method for GGM evaluations across a number of metrics,
    \item We conducted systematic experiments under different settings to evaluate the fidelity and diversity of graph generative models,
    \item Our experimental results demonstrate the superiority of our method across various metrics compared to other deep learning-based baselines.
    \item Although our method excels across a number of metrics, no single method stands out across all metrics and datasets. Therefore, this work also emphasizes the need for further research into evaluation techniques for graph generative models.
\end{enumerate}

In the following sections, we first provide a brief review of related works in Section \ref{sec:related}. We then introduce the necessary background knowledge in Section \ref{sec:pre} to facilitate the understanding of our work. Next, we delve into our proposed method, explaining how to leverage masked autoencoders to evaluate graph generative models in Section \ref{sec:gmae}. Subsequently, we present the experimental setup and results in Section \ref{sec:exp}. Finally, we conclude our work in Section \ref{sec:conc}.

\section{Related Work}\label{sec:related}
\subsection{Graph Generative Models} In recent years, researchers have proposed various types of graph generative models \cite{ggm_survey_zhu2022a}. \cite{graphRNN} proposed GraphRNN, an auto-regressive model, which factorizes the generation process in a sequential way. \cite{simonovsky2018graphvae} proposed GraphVAE, which maximizes a Evidence Lower Bound in an encoder-decoder structure. Normalizing flow models have also been applied to graph generation \cite{graph_normalizing_flow_graphDF}, which estimates the density of graph distributions via the change of variable theorem. \cite{de2018molgan} proposed GAN-based graph generative models. The latest diffusion generative models were also adapted to graph domain \cite{pmlr-v108-niu20a-diffusion-GGM}.

Note that although the proposed method in our paper is designed to evaluate graph generative models, it is \emph{not restricted to any specific type of GGMs}. This is because it assesses the graphs generated by these models, rather than directly on the generative models themselves. The generated graphs are obtained by perturbing real graph datasets, simulating a graph generative model that generates graphs with a distribution different from the training graphs. Therefore the proposed method is both \emph{model-agnostic} and \emph{application-agnostic}.

\subsection{Metrics for Graph Generative Models Evaluation}
The metrics are used to compare generated graphs to real training graphs. They compare both fidelity and diversity. To compare fidelity/similarity, one can utilize Fréchet Inception Distance (FID) \cite{FID} or MMD \cite{MMD} with different kernels. To assess diversity, one needs check if there is mode collapse or mode dropping. These can be examined with Improved Precision \& Recall \cite{recall} or Density and Coverage \cite{coverage} metrics.

To extract graph representations, traditionally one uses statistics from the node degree distribution, clustering coefficients or Laplacian spectrum distributions \cite{o'bray2022evaluation}. More recently, \cite{thompson2022evaluation} proposed a deep learning-based method for graph generative models evaluation, where they employ random graph neural networks (GNN) to extract graph features. Building upon this, a follow-up paper \cite{ggm_shirzad2022evaluating} proposed to contrastively learn graph features, which generally performs better according to their experimental results.

\subsection{Graph Masked Autoencoder}
\cite{he2022masked} proposed masked autoencoder in vision domain, which is a self-supervised learning method that masks random patches of the input image and then reconstructs them with learnable neural networks. \cite{graphmae_mask} adapted the method to the graph domain, masking node attributes only. \cite{maskgae_JintangLi} is another graph mask autoencoder paper, which masks edges or paths. In this paper, we perturb either nodes or edges for each graph, with a predefined probability, as discussed in Section \ref{subsec:eval_setup}.

\section{Preliminaries}\label{sec:pre}
In this section, we introduce background knowledge to facilitate the understanding of this paper. To evaluate a graph generative model, we need to compare the graphs generated by the GGM against the real training graphs with various metrics to evaluate the fidelity and the diversity of the GGM.

To evaluate the fidelity of a GGM, one needs to measure how similar the generated graphs are to the real graphs. In this section, we discuss some of the commonly used metrics.

\subsection{Fréchet Distance (FD)} It assumes the real graphs and the generated graphs are two multivariate Gaussian distributions with mean $\mu$ and covariance $\mathbf{C}$. It compares the mean and variance of two distributions. More specifically, the difference of the two distributions are computed by the distance \cite{FID}
\begin{equation}
    \text{FD}(\mathbb{H}_r, \mathbb{H}_g) = \|\mathbf{\mu}_r - \mathbf{\mu}_g\| + \text{Tr}(\mathbf{C}_r + \mathbf{C}_g - 2(\mathbf{C}_r \mathbf{C}_g)^{1/2})
\end{equation}
where $\mathbb{H}$ represents the distributions of graph representations.

\subsection{Maximum Mean Discrepancy (MMD)} This metric measures the dissimilarity of two distributions $H_r$ and $H_g$ and a lower value of MMD means the two distributions are closer \cite{MMD,thompson2022evaluation}.
\begin{multline}
\text{MMD}(\mathbb{H}_g, \mathbb{H}_r) := \frac{1}{m^2} \sum_{i,j=1}^{m} k(x_i^r, x_j^r) + \frac{1}{n^2} \sum_{i,j=1}^{n} k(x_i^g, x_j^g) \\
- \frac{2}{nm} \sum_{i=1}^{n} \sum_{j=1}^{m} k(x_i^g, x_j^r)
\end{multline}
where $k(\cdot,\cdot)$ is a kernel function. In this paper, we use two kernels: RBF kernel $k(x_i, x_j)=\text{exp}(-\frac{d(x_i,x_j)}{2\sigma^2})$ and linear kernel $k(x_i,x_j)=x_i^T\cdot x_j$ \cite{ggm_shirzad2022evaluating}.

To evaluate the diversity of a GGM, one needs to check the two possible issues of a flawed generative models: mode dropping and mode collapse \cite{metric-gan,thompson2022evaluation}. Real graphs are usually diverse, but the generated graphs may ignore some modes, causing mode dropping, or lack diversity within some modes, causing mode collapse. The following metrics could be used to measure the diversity.

\subsection{Improved Precision \& Recall (P\&R)} This metric \cite{recall} constructs the manifold separately for both real graphs and generated graphs by extending a radius from each sample to its k-th nearest neighbors to form a hypersphere and then take the union of all the hyperspheres. The precision measure reflects the probability of generated samples falling within the manifold of real samples, while the recall measure reflects the probability of real samples falling within the manifold of generated samples.

\subsection{Density \& Coverage (D\&C)} This is introduced as a more robust metric \cite{coverage} than Precision \& Recall. It is based on the concepts of P\&R. The density measure counts how many real sample hyperspheres contain a generated sample instead of just checking if the generated sample is contained within any real sample neighborhood hypersphere. The coverage measure is similar to recall by counting the ratio of real samples covered by generated ones, but it builds the nearest neighbor hypersphere around the real samples instead of the generated ones. The harmonic mean F1 of P\&R and D\&C are both used in this paper.

\begin{figure*}[h]
     \hspace{-8mm}
     \includegraphics[width=1.1\textwidth,left]{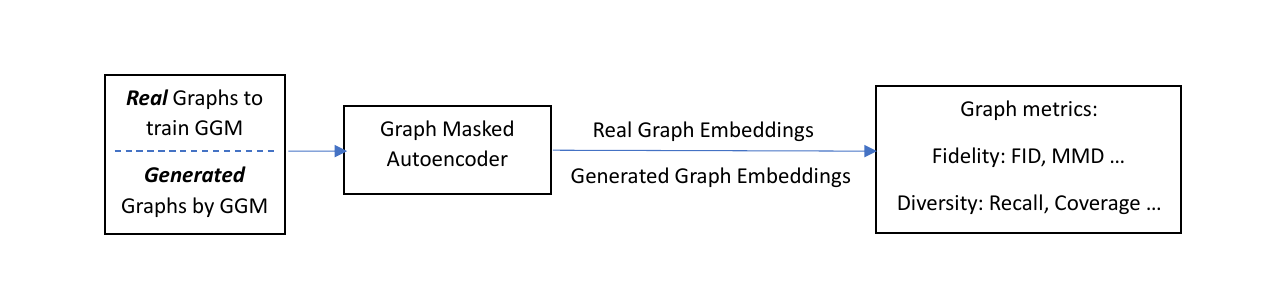}
     \caption{The Process to Evaluate Graph Generative Models with Graph Masked Autoencoder}
     \label{fig:process}
 \end{figure*}

\section{Graph Masked Autoencoder for Evaluation}\label{sec:gmae}
In this paper, we propose to employ Graph Masked Autoencoder (GMAE) to extract graph features for further GGM evaluations. The whole process for GGM evaluation is shown in Figure \ref{fig:process}.
We utilize the architecture from \cite{graphmae_mask} for node feature masking and the architecture from \cite{maskgae_JintangLi} for edge masking. The masking architecture learns graph representations by recovering the masked node features or edges. Except for the GMAE training process, the rest of the evaluation process is similar to those in \cite{thompson2022evaluation,ggm_shirzad2022evaluating} and we largely follow their process for a fair comparison of the experimental results. Note that the primary function of GMAE is to extract accurate graph representations, a crucial step in the GGM evaluation process.

For a proposed GGM evaluation technique to be effective, it must distinguish the distribution difference between real graphs and generated graphs, and reflect any mode dropping or collapse occurred in the generated graphs. In the experiments, the \emph{real} graphs refer to the datasets described in Section \ref{subsec:datasets}. The \emph{generated} graphs are obtained by perturbing the real graphs. That is, the ``Generated Graphs by GGM" in Figure \ref{fig:process} result from this perturbation, highlighting why the evaluation is model-agnostic. Greater perturbation of the real graphs results in a larger distribution difference between the real and the generated graph set. A robust GGM evaluation technique should accurately differentiate the degree of variation between them.

To assess the GGM evaluation techniques, we follow the approach in \cite{thompson2022evaluation} by monotonically increasing the degree $t$ of real graphs perturbation, then compare the set of the perturbed graphs with the set of the original real graphs using the metrics mentioned in Section \ref{sec:pre}. For each degree of perturbation $t\in[0,1]$, we can get a normalized metric score $\hat{s}$. For a strong GGM evaluation technique, ideally the metric score $\hat{s}=0$ if the two set of graphs are the same (i.e., when degree of perturbation $t=0$), $\hat{s}=1$ if the degree of perturbation $t=1$ and $\hat{s}$ monotonically increasing as degree of perturbation $t$ increases from 0 to 1. This relationship between $t$ and $\hat{s}$ can be captured by the Spearman's rank correlation coefficient. If a GGM evaluation technique faithfully learns graph representations (e.g., with GMAE), the Spearman rank correlation will be $1$, otherwise it will be less than 1 and the worst value is $-1$. Therefore, the Spearman rank correlation is employed to assess whether our proposed method---leveraging GAME---can accurately extract graph representations and, thus, whether it, combined with other components as shown in Figure \ref{fig:process}, constitutes a better GGM evaluation technique. We provide more details about the graph perturbation methods in the experimental Section \ref{sec:exp}.

\begin{table}[h]
\caption{The Statistical Overview of the Datasets}
\label{tab:datasets}
\begin{adjustbox}{width=0.7\textwidth, center} 
\begin{tabular}{|l|l|l|l|}
\hline
                   & REDDIT-MULTI-5K & DBLP\_v1 & Proteins    \\ \hline
num of graphs      & 4410            & 17892    & 739         \\ \hline
mean num nodes     & 378.8           & 11.2     & 52.8        \\ \hline
min num nodes      & 22              & 3        & 20          \\ \hline
max num nodes      & 1000            & 39       & 620         \\ \hline
mean num edges     & 433.4           & 21.3     & 98.7        \\ \hline
min num edges      & 21              & 2        & 23          \\ \hline
max num edges      & 1638            & 168      & 1049        \\ \hline

\end{tabular}
\end{adjustbox}
\end{table}

\section{Experiments}\label{sec:exp}

\subsection{Datasets}\label{subsec:datasets}
In our experiments, we use the datasets REDDIT-MULTI-5K, DBLP\_v1, and Proteins \cite{tudataset}, spanning the domain of social networks and bioinformatics. We remove from the dataset graphs with less than 3 nodes (less than 20 nodes for Proteins) or with more than 1{,}000 nodes. For REDDIT-MULTI-5K and DBLP\_v1 datasets, we randomly sample approximately $800$ to 1{,}000 graphs per run due to computational complexity. These datasets have graph node counts that range from a few to one thousand, and edge counts from two to over one thousand. A statistical overview of these datasets is provided in Table \ref{tab:datasets}.

\subsection{Baselines}\label{sec:baseline} In our experiments, we compare with two deep learning-based GGM evaluation techniques. \cite{thompson2022evaluation} employs random graph neural networks to extract the representations of graphs. In contrast, \cite{ggm_shirzad2022evaluating} leverages a contrastive learning method to extract graph representations. These two baselines serve as the reference points for evaluating the effectiveness of our proposed method. Note that we do not use traditional graph representation-based techniques as baselines due to their limitations as discussed in Section \ref{sec:intro}.

\begin{figure*}[h!]
    \captionsetup[subfigure]{skip=-8pt}
    \centering
    \subcaptionbox{}{\includegraphics[width=0.45\textwidth]{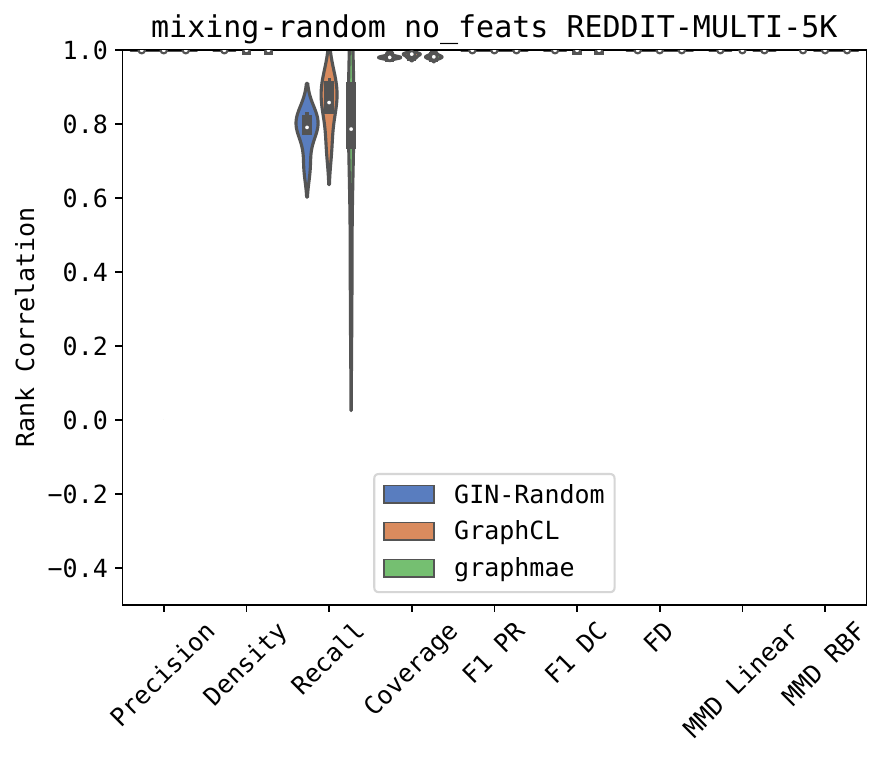}}
    \hspace{0.5cm}
    \subcaptionbox{}{\includegraphics[width=0.45\textwidth]{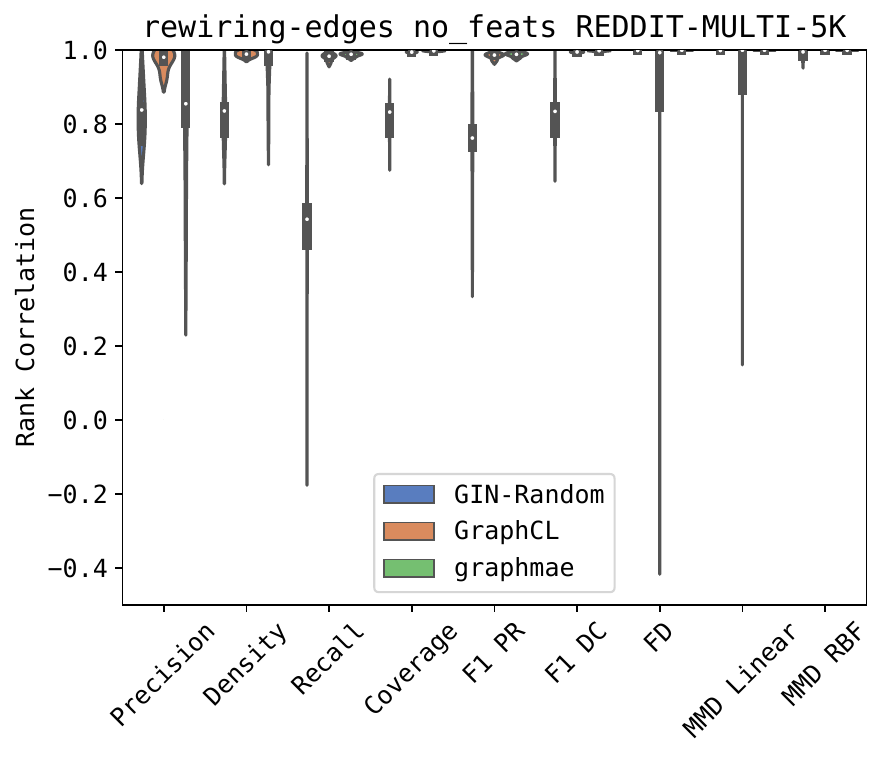}}
    \\
    \subcaptionbox{}{\includegraphics[width=0.45\textwidth]{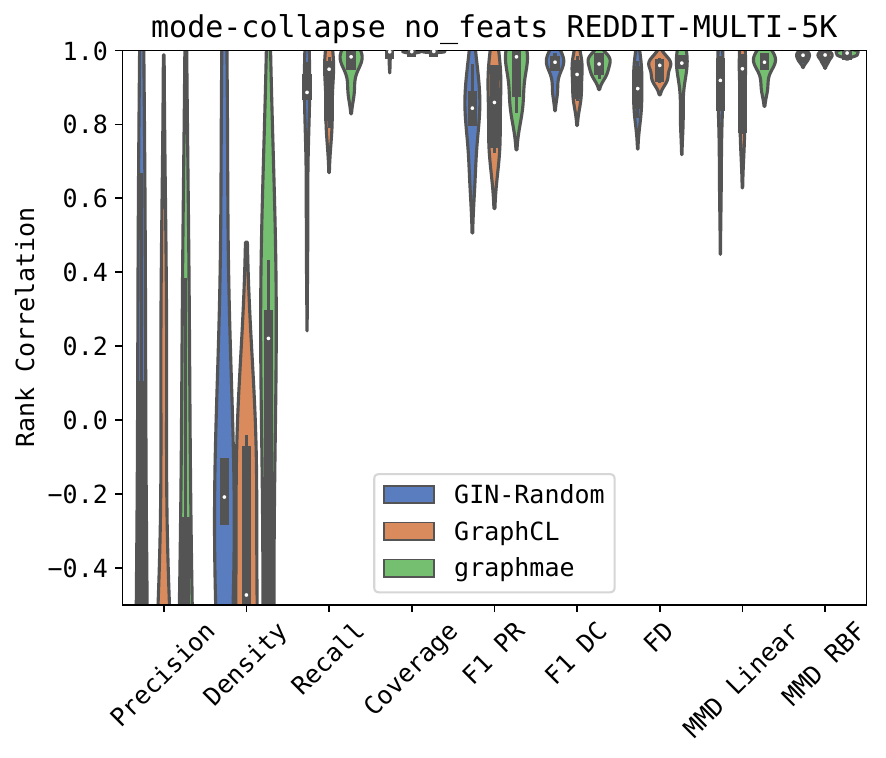}}
    \hspace{0.5cm}
    \subcaptionbox{}{\includegraphics[width=0.45\textwidth]{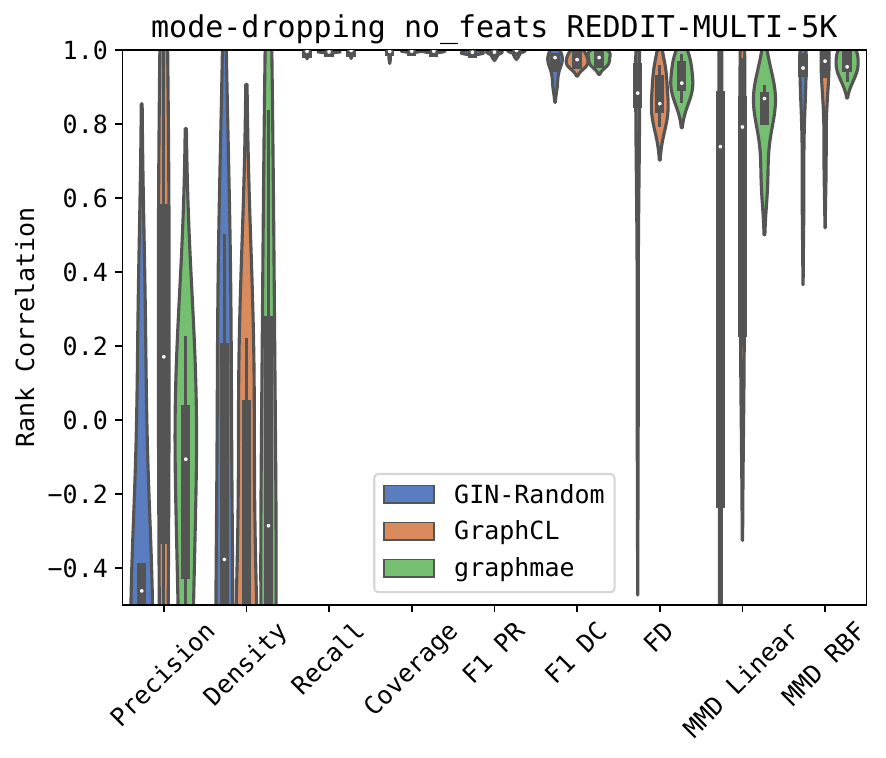}}
    \caption{Experimental results across the perturbation methods for \texttt{REDDIT-MULTI-5K} dataset. A higher and shorter violin plot indicates better results.}
    \label{fig:no_feats_REDDIT-MULTI-5K}
\end{figure*}

\subsection{Fidelity and Diversity Measurement Setup}\label{sec:setup}
We adhere to the procedures outlined in \cite{thompson2022evaluation} to perturb the graphs. The degree of perturbation ranges from 0 to 1, with step size 0.01.

Fidelity measurement involves two types of perturbations: randomly mixing graphs and rewiring edges. To randomly mix graphs, we replace real graphs with Erdős–Rényi (E-R) graphs with the same number of nodes and proportion of edges. To rewire edges, each edge in a graph is rewired with a probability equal to the degree of perturbation. If an edge is rewired, we pick one of its two nodes with equal probability, then disconnect it from the picked node and connect it to a randomly selected node from the graph.

For diversity measurement, we follow \cite{thompson2022evaluation} to simulate mode collapse and mode dropping. Both experiments start with identifying clusters of the real graph set. For mode collapse, we replace the graphs in a cluster with its respective cluster center. For mode dropping, we replace the graphs in the dropped cluster with graphs from the remaining clusters. During the experiments, we gradually increase the number of collapsed or dropped clusters from 0 until it reaches to the total number of clusters.

\subsection{The Evaluation Setup} \label{subsec:eval_setup}
We use the default values in the released code from \cite{ggm_shirzad2022evaluating} for the baselines setting. For each experiment, we run the experiment five times with different seeds, then generate a corresponding violin plot. \textit{The graph masked autoencoder mask either nodes or edges with equal probability}. The mask rate is 0.2 in the experiments.

\begin{figure*}[h!]
    \captionsetup[subfigure]{skip=-8pt}
    \centering
    \subcaptionbox{}{\includegraphics[width=0.45\textwidth]{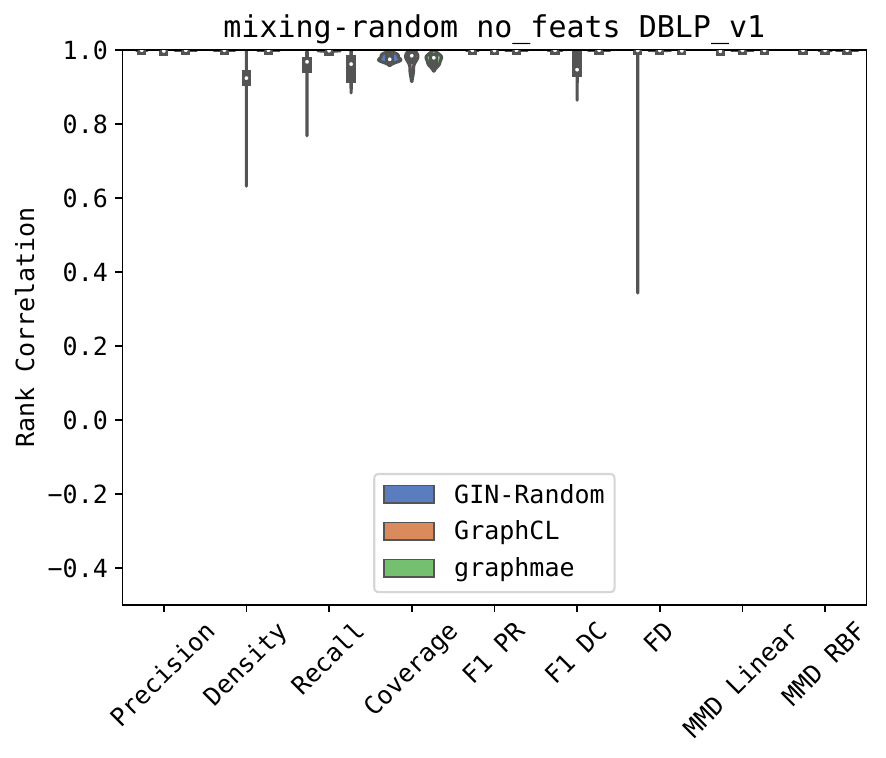}}
    \hspace{0.5cm}
    \subcaptionbox{}{\includegraphics[width=0.45\textwidth]{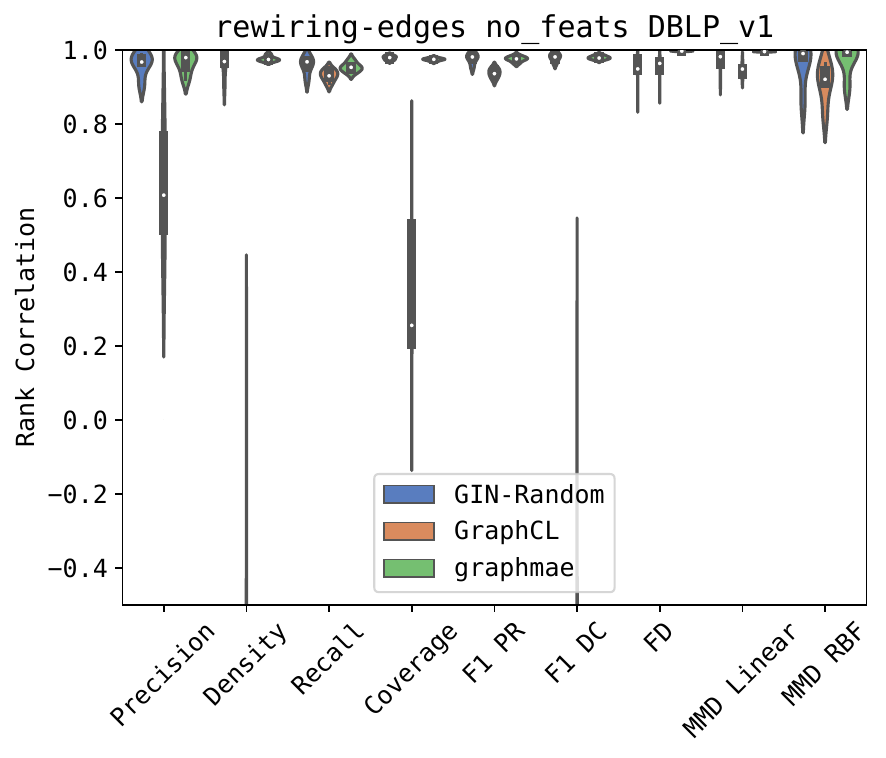}}
    \\
    \subcaptionbox{}{\includegraphics[width=0.45\textwidth]{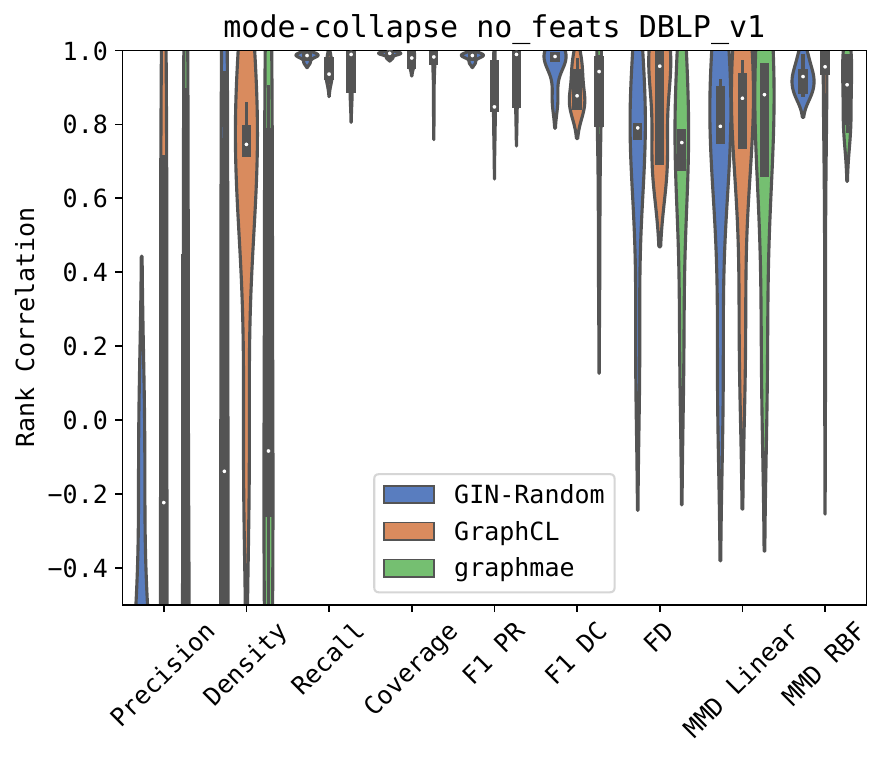}}
    \hspace{0.5cm}
    \subcaptionbox{}{\includegraphics[width=0.45\textwidth]{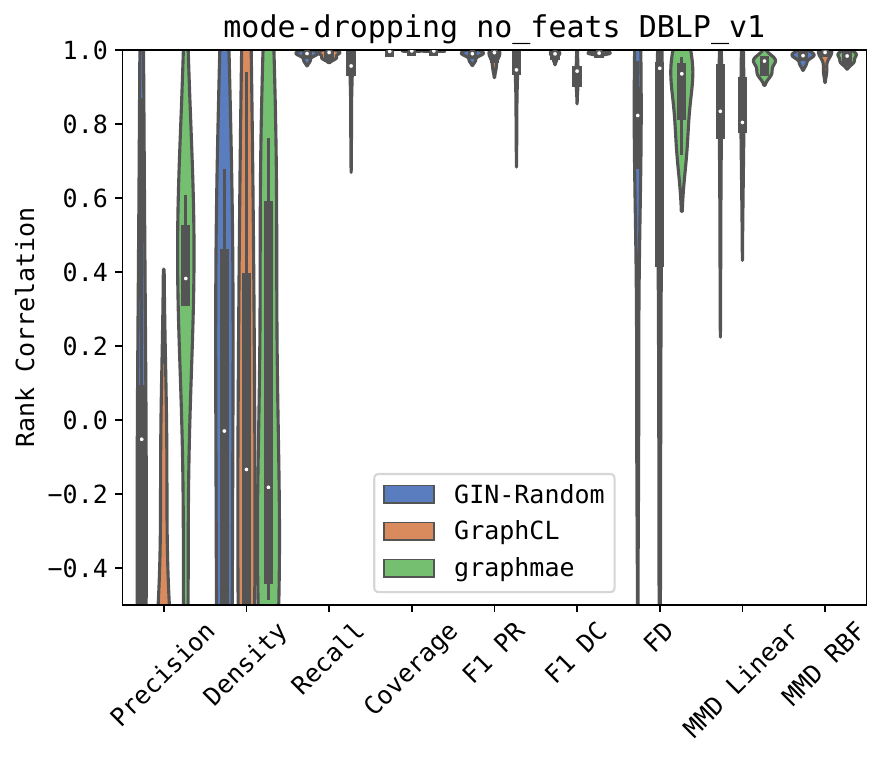}}
    \caption{Experimental results across the perturbation methods for \texttt{DBLP\_v1} dataset. A higher and shorter violin plot indicates better results.}
    \label{fig:no_feats_DBLP_v1}
\end{figure*}

\subsection{Results} \label{sec:results}
\begin{figure*}[h!]
    \captionsetup[subfigure]{skip=-8pt}
    \centering
    \subcaptionbox{}{\includegraphics[width=0.45\textwidth]{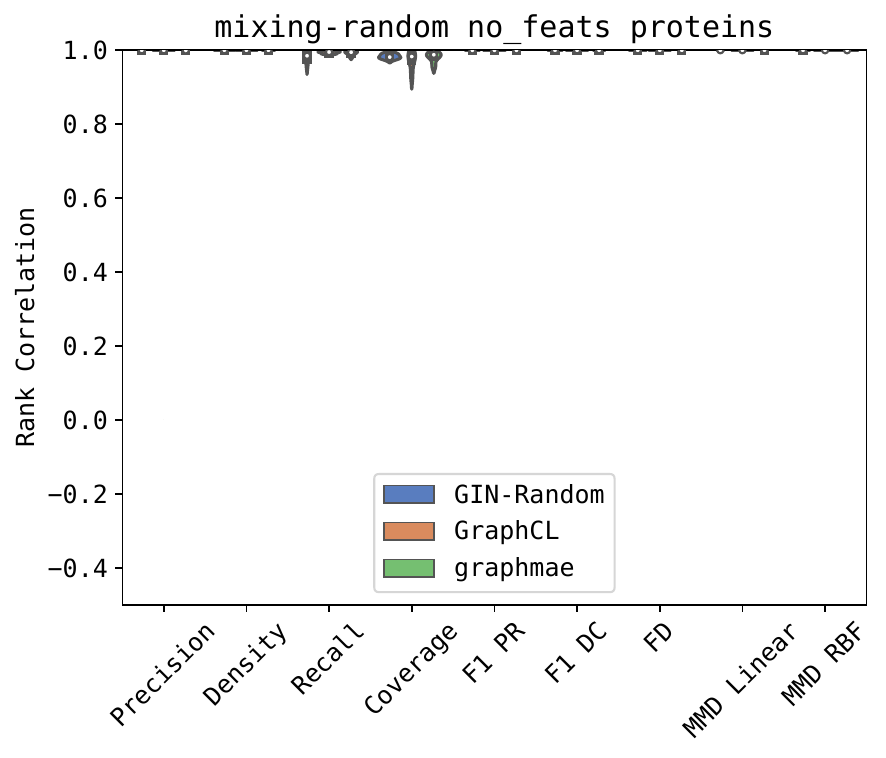}}
    \hspace{0.5cm}
    \subcaptionbox{}{\includegraphics[width=0.45\textwidth]{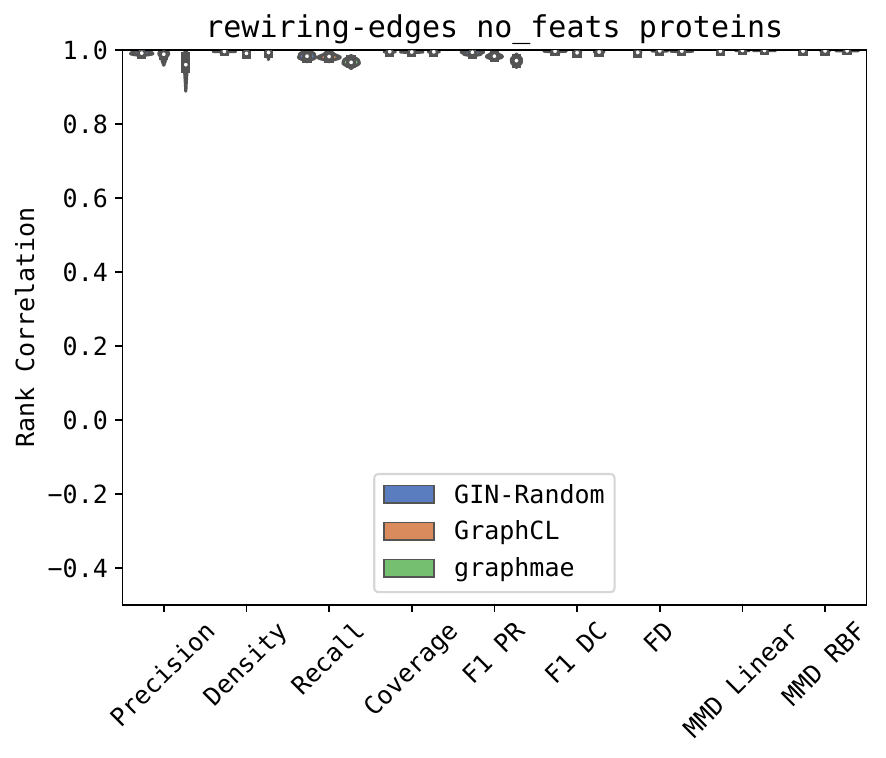}}
    \\
    \subcaptionbox{}{\includegraphics[width=0.45\textwidth]{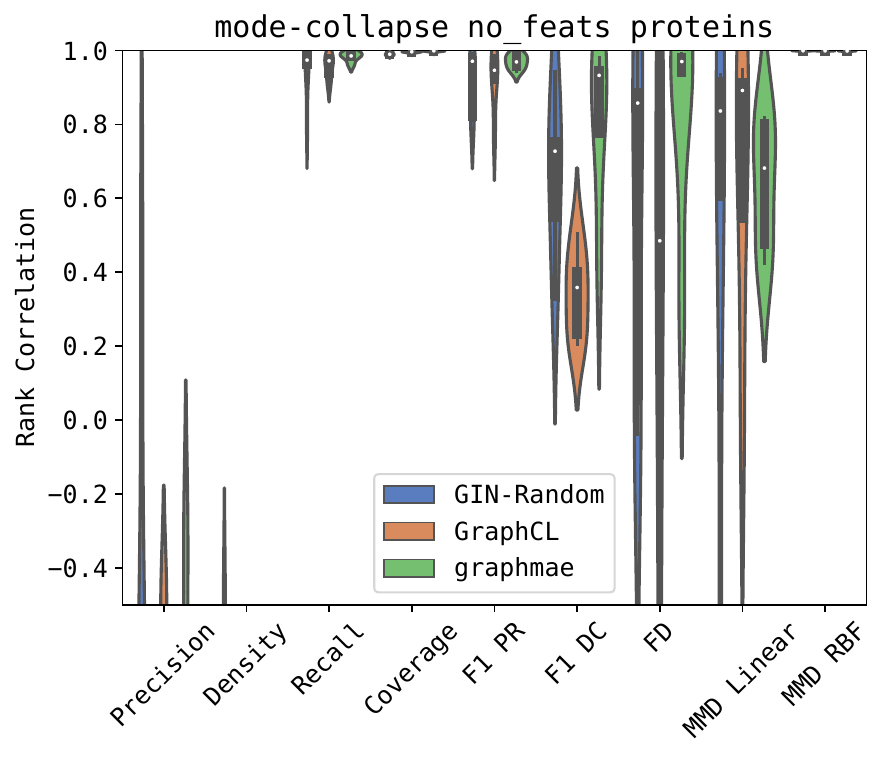}}
    \hspace{0.5cm}
    \subcaptionbox{}{\includegraphics[width=0.45\textwidth]{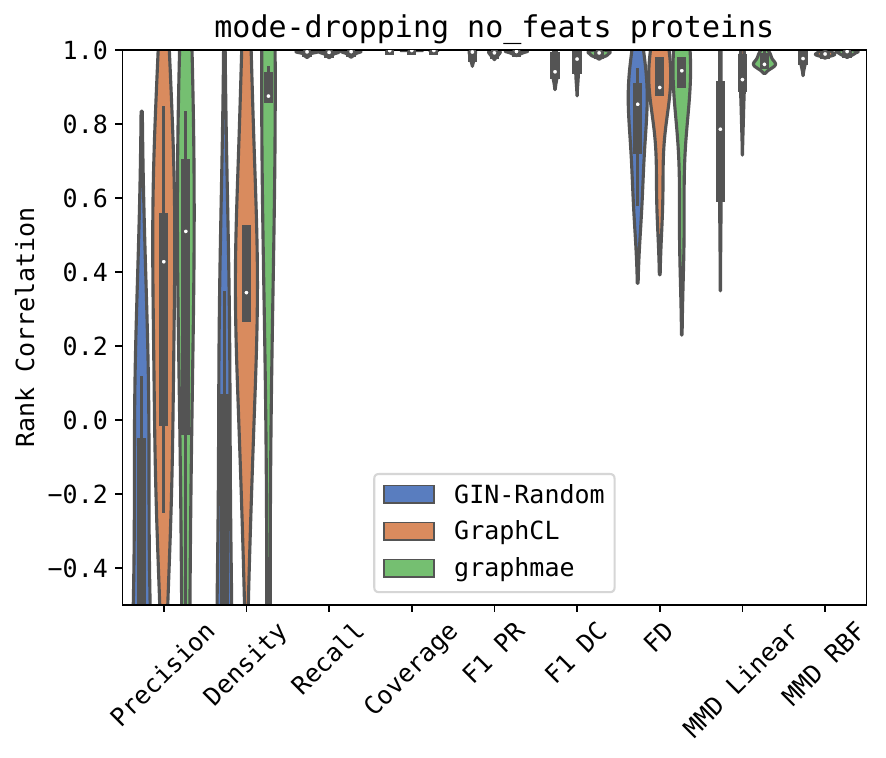}}
    \caption{Experimental results across the perturbation methods for \texttt{proteins} dataset. A higher and shorter violin plot indicates better results.}
    \label{fig:no_feats_proteins}
\end{figure*}

We show the experimental results in Figure \ref{fig:no_feats_REDDIT-MULTI-5K}, \ref{fig:no_feats_DBLP_v1} and \ref{fig:no_feats_proteins} for each dataset respectively. The x-axis of these figures corresponds to the different metrics we discussed in Section \ref{sec:pre}, and the y-axis represents the Spearman correlation mentioned in Section \ref{sec:gmae}. As discussed in Section \ref{sec:gmae}, the ideal Spearman rank correlation is $1$ and the worst one is $-1$. We use violin plots to illustrate the results, visualizing the distribution of the data --- in this case, the \emph{distribution} of the Spearman correlation coefficients.

In the figures above, the white dot in each violin plot represents the \emph{median}, while the thick black bar indicates the interquartile range (IQR), which is the distance between the upper and lower quartiles. The whiskers extend to the farthest data points within 1.5 IQR from the box. The violin shape represents the \emph{probability density} of the data, smoothed by kernel density estimation, with horizontally wider green sections indicating higher probability regions.

In summary, a \emph{higher and shorter} violin plot indicates better results, corresponding to a higher Spearman correlation with lower variance.

The figures indicate that although no method consistently stands out across all metrics, there are a number of metrics for each dataset where our method performs best as detailed below.

Figure \ref{fig:no_feats_REDDIT-MULTI-5K} presents the experimental results for dataset REDDIT-MULTI-5K. Our method performs best across most metrics, achieving higher medians and/or lower variability. For instance, in the mode collapse perturbation, our method excels in metrics ``recall" and ``MMD Linear". Similarly, in mode dropping perturbation, it outperforms the other methods in ``FD", ``MMD Linear" and ``MMD RBF". Notably, under the rewiring edges perturbation, the random GNN baseline performs significantly worse than the other two methods in ``Recall", while the contrastive learning baseline shows the poorest performance in ``FD" and ``MMD Linear".

Figure \ref{fig:no_feats_DBLP_v1} presents the experimental results for dataset DBLP\_v1. In the mode dropping perturbation, our method achieves the best performance in ``FD" and ``MMD Linear", with higher medians and/or lower variability. Notably, under the rewiring edges perturbation, the contrastive learning baseline performs significantly worse than the other two methods across several metrics, such as ``precision", ``density" and ``coverage".

Figure \ref{fig:no_feats_proteins} presents the experimental results for dataset Proteins. Our method and the two baselines have similar performance across most metrics. However, our method stands out with a higher median \textit{and} lower variability for ``FD" in the mode collapse perturbation, and ``MMD linear" in the mode dropping perturbation.

\subsection{Discussions}
Since GGM metrics are usually not interchangeable, and each of them evaluates GGM from its unique perspective, one should consider the dataset and the intended application when choosing the appropriate graph representation learning technique and evaluation metrics.

The reason that our proposed method performs better across a number of metrics is that graph masked autoencoder can perform comparable or better compared to the contrastive learning-based method on graph representation learning, as shown in \cite{graphmae_mask,maskgae_JintangLi}, where they performed graph classification or node classification tasks. Additionally, contrastive learning-based graph representation method performs better than random neural network-based method according to \cite{ggm_shirzad2022evaluating}.
Thus, our method provides another approach for effective graph generative models evaluations, especially when using the metrics discussed above, such as ``FD" and ``MMD Linear".

The performance of our proposed method is expected to improve further by adjusting architectures, training strategy, and hyperparameters of GMAE, but this may not succeed without significant tuning efforts.


While our method generally outperforms others in these experiments, we emphasize that neither our approach nor the baselines consistently excels across all metrics. This underscores the need for further research in GGM evaluation techniques, especially in light of recent advances in generative models.

\section{Conclusion}\label{sec:conc}
In this paper, we propose graph masked autoencoder-based approach to learn graph representations for graph generative models evaluations. Compared to random GNN and contrastive graph learning-based approaches, the experimental results demonstrate that our method performs better across a number of metrics, such as ``recall", ``FD" and ``MMD Linear" as discussed in Section \ref{sec:results}. Our findings suggest that when evaluating generative graph models, one should consider multiple deep learning-based graph representation learning methods for a more comprehensive evaluation. Additionally, our work highlights the significance and challenges associated with evaluating graph generative models in practice.


\bibliographystyle{plain}
\bibliography{references}


\end{document}